\documentclass[letterpaper]{article} 
\usepackage[]{aaai23}  
\usepackage{times}  
\usepackage{helvet}  
\usepackage{courier}  
\usepackage[hyphens]{url}  
\usepackage{graphicx} 
\usepackage{natbib}  
\usepackage{caption} 
\usepackage{algorithm}
\usepackage{algorithmic}

\usepackage{newfloat}
\usepackage{listings}
\DeclareCaptionStyle{ruled}{labelfont=normalfont,labelsep=colon,strut=off} 
\floatstyle{ruled}
\newfloat{listing}{tb}{lst}{}
\floatname{listing}{Listing}
\usepackage{graphicx}
\usepackage{amsmath}
\usepackage{amssymb}
\usepackage{booktabs}

\usepackage{times}
\usepackage{epsfig}
\usepackage{graphicx}
\usepackage{amsmath}
\usepackage{amssymb}

\usepackage[utf8]{inputenc} 
\usepackage[T1]{fontenc}    
\usepackage{url}            
\usepackage{booktabs}       
\usepackage{amsfonts}       
\usepackage{nicefrac}       
\usepackage{microtype}      
\usepackage{xcolor}         

\usepackage{booktabs}                                   
\usepackage{multirow}                                   
\usepackage{tablefootnote}                              
\usepackage[symbol]{footmisc}                           

\usepackage{color}                                      
\usepackage{pifont}
\usepackage{bm}

\usepackage[normalem]{ulem}                             

\newcommand{\mR}        {\mathcal{R}}
\newcommand{\mM}        {\mathcal{M}}
\newcommand{\mA}        {\mathcal{A}}

\newcommand{\bF}        {\textbf{F}}
\newcommand{\bbf}        {\textbf{f}}
\newcommand{\bV}        {\textbf{V}}
\newcommand{\bQ}        {\textbf{Q}}
\newcommand{\bC}        {\textbf{C}}
\newcommand{\bv}        {\textbf{v}}
\newcommand{\bq}        {\textbf{q}}
\newcommand{\bc}        {\textbf{c}}
\newcommand{\btheta}    {\bm{\theta}}
\newcommand{\bG}        {\textbf{G}}
\newcommand{\bW}        {\textbf{W}}
\newcommand{\bM}        {\textbf{M}}
\newcommand{\bh}        {\textbf{h}}
\newcommand{\bb}        {\textbf{b}}
\newcommand{\batt}        {\textbf{att}}
\newcommand{\balpha}        {\bm{\alpha}}
\newcommand{\bx}        {\textbf{x}}
\newcommand{\bs}        {\textbf{s}}
\newcommand{\bp}        {\textbf{p}}

\newcommand{\para}[1]{\noindent\textbf{#1}.\ }

\newcommand{\look}[2]{{\color{#1}#2}}                   

\newcommand{\zp}[1]{\look{magenta}{#1}}      

\newcommand{\eat}[1]{}                                  

\newcommand{\task}[1]   {\textit{#1}}

\newcommand{\tframeqa}  {\task{FrameQA}}
\newcommand{\tcount}  {\task{Count}}
\newcommand{\taction}  {\task{Action}}
\newcommand{\ttrans}  {\task{Trans}}

\newcommand{\twhat}  {\task{What}}
\newcommand{\twho}  {\task{Who}}
\newcommand{\thow}  {\task{How}}
\newcommand{\twhen}  {\task{When}}
\newcommand{\twhere}  {\task{Where}}
\newcommand{\tall}  {\task{All}}

\def\Ours{EC-GNNs}

\title{Cross-Modal Reasoning with Event Correlation for Video Question Answering}

\author{Chengxiang Yin$^{1}$, Zhengping Che$^{2}$, Kun Wu$^{1}$, Zhiyuan Xu$^{2}$, Qinru Qiu$^{1}$, Jian Tang$^{2}$\\
$^{1}$Syracuse University, $^{2}$Midea Group \\
}

\begin{document}

\maketitle

\begin{abstract}
%
%
Video Question Answering (VideoQA) is
an attractive
and challenging research direction aiming to
understand complex semantics of
heterogeneous data from two domains, i.e., the spatio-temporal video content and the word sequence in question.
%
%
%
Although various attention mechanisms have been utilized to manage contextualized representations by modeling intra- and inter-modal relationships of the two modalities, one limitation of the predominant VideoQA methods is the lack of reasoning with event correlation, that is, sensing and analyzing relationships among abundant and informative events contained in the video.
%
%
In this paper, we introduce the dense caption modality as a new auxiliary and distill event-correlated information from it to infer the correct answer.
%
%
To this end, we propose a novel end-to-end trainable model, \textbf{E}vent-\textbf{C}orrelated \textbf{G}raph \textbf{N}eural \textbf{N}etwork\textbf{s} (\textbf{EC-GNNs}), to perform cross-modal reasoning over information from the three modalities (i.e., caption, video, and question).
%
%
%
Besides the exploitation of a brand new modality,
we employ cross-modal reasoning modules for explicitly modeling inter-modal relationships and aggregating relevant information across different modalities, and we propose a question-guided self-adaptive multi-modal fusion module to collect the question-oriented
and event-correlated evidence through multi-step reasoning.
%
%
%
%
%
We evaluate our model on two widely-used benchmark datasets and conduct an ablation study to justify the effectiveness of each proposed component.
%
%
\end{abstract}

\section{Introduction}
\begin{figure}[t!]
\centering
\includegraphics[width=0.99\columnwidth]{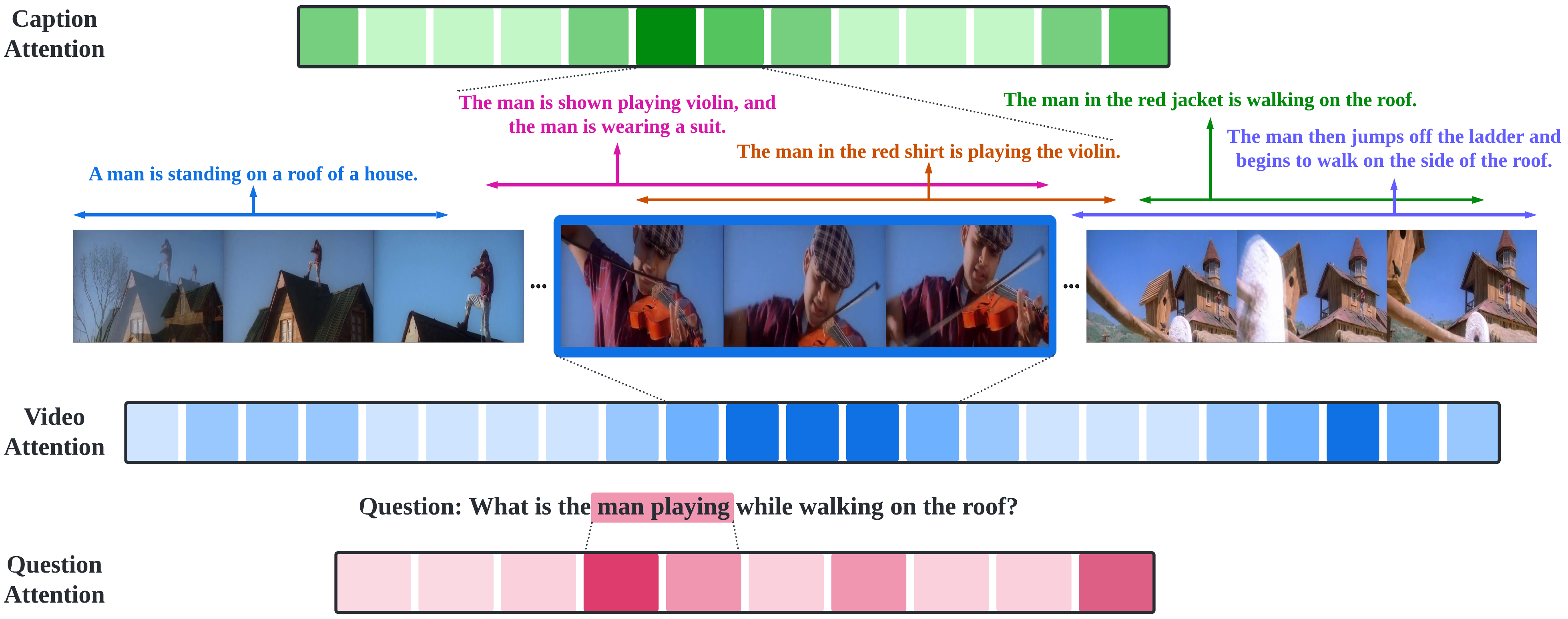}
\caption{
An illustration of how {\Ours} handle VideoQA tasks.
{\Ours} aim to associate semantic contents in the generated dense captions and relevant visual contents in frame sequence with the actual subject queried in question.
For a complex question such as ``What is the man playing while walking on the roof?'', {\Ours} first localize when the man is walking on the roof and then focus on what the man is playing (i.e., the violin).}
\label{Fig::Motivation}
\end{figure}

Video Question Answering~(VideoQA)~\cite{zhu2017uncovering,zeng2017leveraging,jang2019video,fan2019heterogeneous,jiang2020reasoning,le2020hierarchical} is a challenging \zp{\eat{computer vision}} task. It can be seen as a proxy task for evaluating a vision-language system's capacity for a more profound understanding of heterogeneous data from two domains, i.e., spatio-temporal video contents and word sequence in question.
A mature VideoQA agent is able to deliver the correct answer by associating relevant visual contents in the video with the actual subject queried in question.
To achieve this, predominant VideoQA methods~\cite{ye2017video,xu2017video,jang2017tgif,jang2019video,li2019beyond} utilize various attention mechanisms to find relevant details in spatial or/and temporal dimensions.
However, none of the existing approaches reasons with events and their correlations for VideoQA.
Temporally localized and occurring concurrently or successively, events are numerous and ubiquitous in video and imply massive knowledge of the video.
For example, in Figure~\ref{Fig::Motivation}, \textit{a man in the red shirt is playing the violin} and then \textit{the man jumps off the ladder and begins to walk on the side of the roof} are two consecutive events depicting the same person.
Sensing such events and analyzing their relationships with video contents and questions provides a promising way to deliver an accurate and explainable answer for a given question.

\eat{Understanding the events involved is vital to spatio-temporal modeling tasks, especially VideoQA.}
Events can be properly understood through modeling the complex video dynamics into a natural language description, namely \textit{video captioning}.
Massive works~\cite{venugopalan2014translating,venugopalan2015sequence,yao2015describing,pan2016jointly} explain video with a single sentence.
For example, they would most likely to focus on \textit{a man in the red shirt is playing the violin} in Figure~\ref{Fig::Motivation}, which provides some details about the man and his outfits but fails to articulate other events in the video.
\eat{While this one-sentence description provides some details about who is playing the violin and depicts its outfits, it fails to recognize and articulate all the other events in the video.}
Instead, incorporating dense video captions~\cite{krishna2017dense,zhou2018end,wang2020dense} as a new auxiliary modality helps to identify and understand more events involved, which provides valuable clues to infer correct answers.

With multiple modalities available in VideoQA, integrating intra- and inter-modal relationships may further benefit the answer inference.
Taking Figure~\ref{Fig::Motivation} as an instance, the model needs to establish subject-predicate relationships between the word \textit{man} and the words \textit{playing} and \textit{walking} in the question.
RNNs are commonly adopted~\cite{zhu2017uncovering,zeng2017leveraging,yin2019memory} but often fail to model the true complexity of language structure especially for semantic-complex questions.
Alternatively, Graph Convolutional Networks (GCNs)~\cite{kipf2016semi} provide a more flexible way to utilize context-aware neural mechanisms to learn the complex intra-modal relationships,
and the requirement of intra-modal modeling and the efficacy of GCNs also hold for videos and dense captions.
One the other hand, the model needs to establish a solid semantic relationship between \textit{the man playing} in the question, the caption \textit{the man in red shirt is playing the violin}, and the few frames highlighted in the middle of the video, to correctly answer the question in Figure~\ref{Fig::Motivation}.
Modeling only one-way interactions~\cite{anderson2018bottom,teney2018tips,jang2017tgif,jang2019video} is not enough,
and methods aggregating relevant information across different modalities and learning long-term semantic dependencies is preferred.
Self-attention~\cite{vaswani2017attention} with different cross-modal key-query pairs provides a promising way of modeling inter-modal relationships.

At the final stage of VideoQA, to aggregate multi-modal features for answer prediction,\eat{To project multi-modal features into a joint space for answer prediction,}
the predominant approaches~\cite{zhu2017uncovering,zeng2017leveraging,yin2019memory} generate a monolithic representation using simple vector operations,
including concatenation, element-wise addition and/or element-wise multiplication.
However, the monolithic representation cannot fully understand cross-modal semantics and often lead to incorrect answers.
As \cite{fan2019heterogeneous} found the sub-optimal performance may come from distraction from the question,
it is promising that multi-step reasoning with question guidance can progressively gather the question-oriented and event-correlated evidences and provide better answers.
\eat{
Inspired by \cite{fan2019heterogeneous},
we design a question-guided self-adaptive multi-modal fusion module, which performs multiple inference loops to collect question-oriented and event-correlated evidence from each modality by progressively refining the joint attention weights over multi-modal features.
with question guidance, the question-oriented and event-correlated evidence can be progressively gathered by performing multiple reasoning steps.
}

In this paper, to deal with VideoQA, we propose \textbf{E}vent-\textbf{C}orrelated \textbf{G}raph \textbf{N}eural \textbf{N}etwork\textbf{s} (\textbf{EC-GNNs}), a novel end-to-end trainable model, which performs reasoning with event correlation, by exploiting the relationships among dense events, video contents and question words.
To the best of our knowledge, this is the first work that explicitly incorporates dense video captions to perform the reasoning of VideoQA.
Specifically, we develop a dense caption modality and handle the VideoQA task via three modality-aware graphs, where the dense caption features, visual features, and question embedding features are kept in the corresponding graphs.
\eat{Specifically, we develop a dense caption modality and handle the VideoQA task via three modality-aware graphs, where the corresponding dense caption features are supported in \textit{caption graph}, appearance and motion features are kept in \textit{video graph}, and the embeddings of question words are provided in \textit{question graph}.}
Three procedures are performed afterwards to infer the final answer.
%
First, graph reasoning modules capture the intra-relations of each modality.
%
Then, cross-modal reasoning modules aggregate relevant information across different modalities by leveraging a cross-modal attention mechanism to model the inter-modal relationships.
%
Finally, after repeating the above two procedures several times, a question-guided self-adaptive multi-modal fusion module conducts multiple inference loops to collect the question-oriented and event-related evidences and refines the final answer prediction.

The contributions are summarized as follows:
1) We propose a novel scheme to perform cross-modal reasoning with event correlation over information from three modalities (i.e., caption, video, and question).
2) We are the first to introduce dense video captions for VideoQA and clarify how to incorporate the event-correlated information into reasoning process.
3)
We explicitly model the inter-modal relationships through a cross-modal attention mechanism in cross-modal reasoning.
4) We design a question-guided self-adaptive multi-modal fusion module that adaptively collects question-oriented and event-correlated evidence.
5) The proposed model outperforms most VideoQA methods and performs on par with the state-of-the-art on two VideoQA benchmarks.

\section{Related Work}
\subsection{Video Question Answering}
%
%
Recently, significant progress has been made to \textit{video question answering (VideoQA)}~\cite{zhu2017uncovering,zeng2017leveraging,jang2017tgif,ye2017video,xu2017video,fan2019heterogeneous,jang2019video,jiang2020reasoning,le2020hierarchical}, which extends \textit{image question answering (ImageQA)}~\cite{antol2015vqa,chen2015abc,anderson2018bottom,teney2018tips,ma2018visual,yin2021hierarchical} to the video domain and requires a more comprehensive spatio-temporal understanding of heterogeneous data from multiple modalities.
As a common practice, most works on VideoQA utilized appearance features and/or motion features to represent video frames
and employ word embedding (e.g., Word2vec~\cite{mikolov2013efficient} and Glove~\cite{pennington2014glove}) to represent question words.
Representative methods~\cite{zhu2017uncovering,zeng2017leveraging,yin2019memory} on VideoQA leveraged RNNs-based encoders to encode frame features and word features into a single representation.
%
Additionally, a number of VideoQA works~\cite{ye2017video,xu2017video,jang2017tgif,jang2019video,li2019beyond} employed temporal attention~\cite{bahdanau2014neural} to attend to the most relevant frames in video.
%
%
%
More recently, some works~\cite{gao2018motion,kim2018multimodal,yin2019memory,fan2019heterogeneous,kim2019progressive} proposed the use of external memory combined with a customized memory addressing mechanism to enhance the spatio-temporal modeling of VideoQA.
%
%
%
%
In this paper, we explore the VideoQA task through new modality generation and cross-modal reasoning with event correlation from three modalities.
%
%

\subsection{Video Captioning and Dense Video Captioning}
%
Similar to VideoQA, the earliest \textit{video captioning} works~\cite{venugopalan2014translating,venugopalan2015sequence,cho2015describing,donahue2015long,pan2016jointly} employed encoder-decoder architectures based on RNNs to model the temporal dynamics of the frame sequences and word sequences.
%
%
Later works~\cite{yao2015describing,gao2017video,li2017gla} applied customized attention mechanisms to selectively focus on salient frames in the video, resulting in better performance.
However, describing video semantics in one sentence often fails to recognize or articulate the other events in the video, which brings the demand of \textit{dense video captioning}~\cite{yu2016video,krishna2017dense,zhou2018end,wang2018bidirectional,wang2021end}.
By extending video captions in the event dimension, dense video captioning model first locates multiple events occurring in the video and generates a
caption for each of them.
%
%
%
A seminal work~\cite{krishna2017dense} employed an existing proposal module~\cite{escorcia2016daps} to detect the event locations and caption each event by utilizing the context from surrounding events.
%
%
%
%
%
As dense events provide valuable clues to infer the correct answer, we explicitly extract and incorporate dense video captions as a new modality for VideoQA reasoning.


\section{Methodology}
\begin{figure*}[t]
\centering
\includegraphics[width=0.99\linewidth]{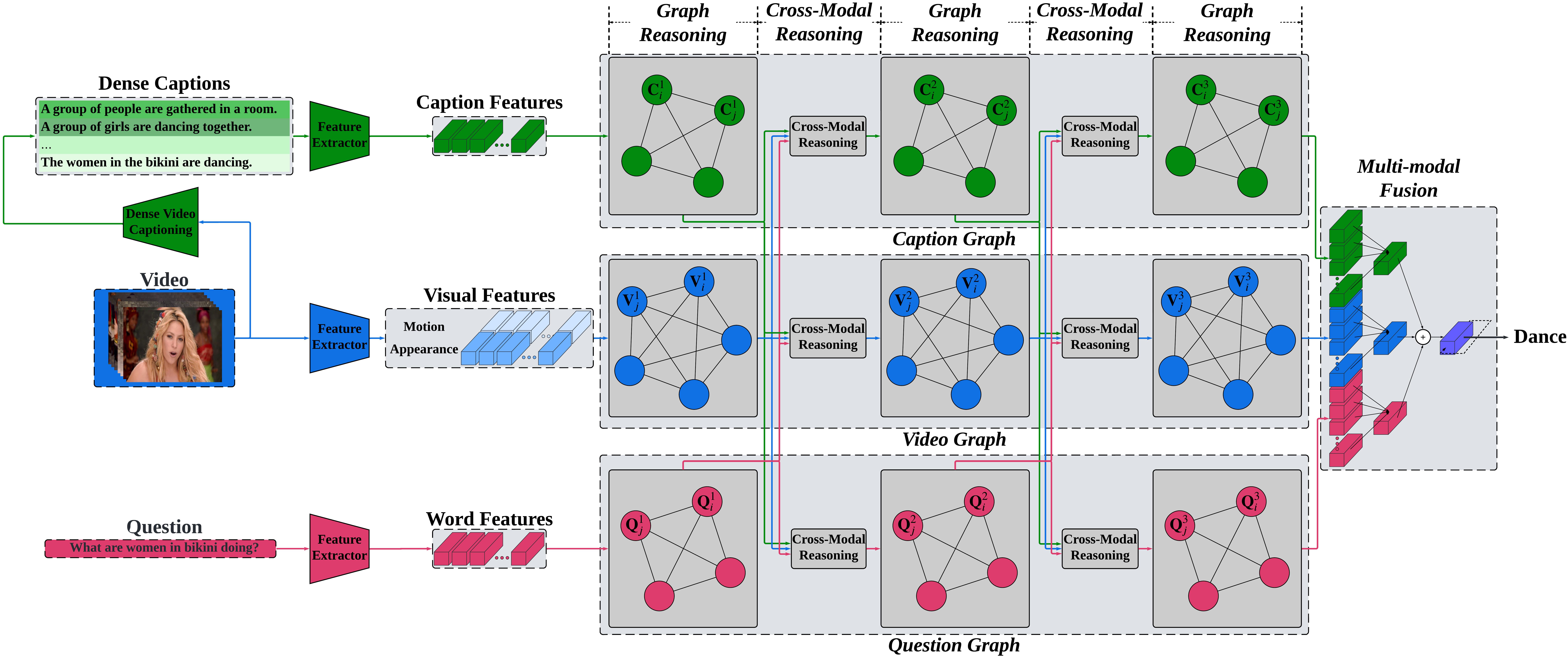}
\caption{The architecture of Event-Correlated Graph Neural Networks (EC-GNNs), which consists of three components: graph reasoning modules, cross-modal reasoning modules, and a question-guided self-adaptive multi-modal fusion module.
%
%
Caption features, visual features, and word features are forwarded to \textit{caption graph}, \textit{video graph}, and \textit{word graph} respectively.
Each of the graphs contains three graph reasoning procedures where two cross-modal reasoning procedures are interleaved.
Finally, multi-modal fusion is performed to infer the answer.
%
%
Green, blue, and red arrows between modules corresponds to the information flow of caption, video, and question modalities respectively.
}
\label{Fig::Architecture}
\vspace{-5pt}
\end{figure*}

This section describes the architecture of the Event-Correlated Graph Neural Networks (EC-GNNs) as shown in Figure~\ref{Fig::Architecture}.
We first introduce the representation~(Sec.~\ref{sec:method-representation}) and graph construction~(Sec.~\ref{sec:method-construction}) of the three modalities (i.e., the caption, the video, and the question).
Then we demonstrate how the graph reasoning modules~(Sec.~\ref{sec:method-graph-reasoning}) and the cross-modal reasoning modules~(Sec.~\ref{sec:method-cross-reasoning}) can model the intra- and inter-relationships.
Finally, we elaborate on the design of the question-guided self-adaptive multi-modal fusion module~(Sec.~\ref{sec:method-fusion}) and the answer prediction~(Sec.~\ref{sec:method-question}).
\subsection{Contextual Representation with Generated Modality}
\label{sec:method-representation}
\para{Generating caption modality}
Dense video captions are a set of captions in descriptive natural language, which describe multiple events temporally localized in a video.
To generate dense captions of a video, we first extract its appearance features and optical flow features using ResNet-200~\cite{he2016deep} and BN-Inception~\cite{ioffe2015batch} respectively.
%
Given extracted features,
we employ the dense video captioning model~\cite{zhou2018end} trained on both ActivityNet Captions~\cite{krishna2017dense} and YouCookII~\cite{zhou2018towards} datasets to obtain the corresponding dense captions.
Each caption is first represented as a sequence of word embeddings based on GloVe 300-D~\cite{pennington2014glove}, and then encoded with a trainable GRU~\cite{chung2014empirical} to obtain its final hidden state as a monolithic representation.
Therefore, the dense captions are denoted as a sequence of monolithic representations ${\bF}^{c} = \{{\bbf}_{i}^{c}: i \leq N_{c}, {\bbf}_{i}^{c} \in {\mR}^{d_{c}}\}$, where $N_{c}$ represents the number of captions and $d_{c}$ is the dimensionality of the caption features.
In Figure~\ref{Fig::Architecture}, we highlight the caption features \textit{in green}.
In order to match the words in captions with the words in questions, same dictionary are used when dealing with dense captions and questions.
%

\para{Representing video and question modalities}
Besides the generated caption modality, we also model directly the video and question modalities.
We extract frame-level appearance features ${\bF}^{a} = \{{\bbf}_{i}^{a}: i \leq N_{v}, {\bbf}_{i}^{a} \in {\mR}^{d_{a}}\}$,
shot-level motion features ${\bF}^{m} = \{{\bbf}_{i}^{m}: i \leq N_{v}, {\bbf}_{i}^{m} \in {\mR}^{d_{m}}\}$,
and common visual features ${\bF}^{v} = \{{\bbf}_{i}^{v}: i \leq N_{v}, {\bbf}_{i}^{v} \in {\mR}^{d_{v}}\}$ from videos,
as well as a sequence of word embeddings ${\bF}^{q} = \{{\bbf}_{i}^{q}: i \leq N_{q}, {\bbf}_{i}^{q} \in {\mR}^{d_{q}}\}$ from questions.
Here, $N_{v}$ is the number of frames in the video, $N_{q}$ is the number of words in the question,
and $d_{a}$, $d_{m}$, $d_{v}$, and $d_{q}$ are dimensionalities of the feature views.
In Figure~\ref{Fig::Architecture}, we highlight the appearance/motion/visual features in \textit{light/normal/dark blue} and the word features in \textit{red}.
More details are provided in Appendix.
%

\para{Obtaining contextualized representations}
All the features of the three modalities (i.e., caption features, visual features, and word features) are time series.
%
%
To exploit the dynamic temporal information and obtain contextual representations for each modality, we leverage three independent GRUs to encode these features separately.
%
\begin{align}
{\bC}^{1}, {\bc}_{N_{c}} &= \mbox{GRU}({\bF}^{c}; {\btheta}_{GRU}^{c})\\
{\bV}^{1}, {\bv}_{N_{v}} &= \mbox{GRU}({\bF}^{v}; {\btheta}_{GRU}^{v})\\
{\bQ}^{1}, {\bq}_{N_{q}} &= \mbox{GRU}({\bF}^{q}; {\btheta}_{GRU}^{q})
\end{align}
The contextualized caption features are denoted as ${\bC}^{1} = \{{\bC}^{1}_{i}: i \leq N_{c}, {\bC}^{1}_{i} \in {\mR}^{d_{C}}\}$,
the contextualized visual features are denoted as ${\bV}^{1} = \{{\bV}^{1}_{i}: i \leq N_{v}, {\bV}^{1}_{i} \in {\mR}^{d_{V}}\}$,
and the contextualized word features are denoted as ${\bQ}^{1} = \{{\bQ}^{1}_{i}: i \leq N_{q}, {\bQ}^{1}_{i} \in {\mR}^{d_{Q}}\}$,
where ${d_{C}}$, ${d_{V}}$, and ${d_{Q}}$ are dimensionalities of the contextualized features;
%
${\bc}_{N_{c}} \in {\mR}^{d_{C}}$, ${\bv}_{N_{v}} \in {\mR}^{d_{V}}$ and ${\bq}_{N_{q}} \in {\mR}^{d_{Q}}$ are the last hidden states, which represent the global features of the three modalities.
%

\subsection{Graph Construction}
\label{sec:method-construction}
Given the contextualized features of three modalities, we construct three modality-aware graphs: \textit{caption graph}, \textit{video graph}, and \textit{question graph}.
As shown in Figure~\ref{Fig::Architecture}, each modality-aware graph is a three-layer fully-connected GNNs (i.e., $l = 1, 2, 3$),
where the caption features ${\bC}^{l}$, visual features ${\bV}^{l}$, and word features ${\bQ}^{l}$ are kept in the \textit{caption}, \textit{video}, \textit{question graphs}, with green, blue and red nodes, respectively.
%
The numbers of nodes in each layer of the three graphs are $N_{c}$, $N_{v}$, and $N_{q}$, respectively,
which equals to the numbers of dense captions, frames, and words in questions.
%
%
%

\subsection{Graph Reasoning}
\label{sec:method-graph-reasoning}
As each layer of the modality-aware GCNs contains modality-specific knowledge, we first capture the intra-modal relationships from each graph by performing graph reasoning independently.
%
%
Given the node features ${\bC}^{l}$ (or ${\bV}^{l}$, ${\bQ}^{l}$),
we measure the semantic correlations among nodes~\cite{wang2018videos}
and obtain the normalized adjacency matrix ${\bG}_{C}^{l} \in {\mR}^{N_{c} \times N_{c}}$
(or ${\bG}_{V}^{l} \in {\mR}^{N_{v} \times N_{v}}$, ${\bG}_{Q}^{l} \in {\mR}^{N_{q} \times N_{q}}$),
and we apply GCNs~\cite{kipf2016semi} to update each node based on its neighbors and the corresponding weights.
Detailed intra-modal graph reasoning procedures are described in Appendix.
%
By doing so, we obtain the updated node features ${\hat{\bC}}^{l}$, ${\hat{\bV}}^{l}$, and ${\hat{\bQ}}^{l}$ accordingly,
which are forwarded to cross-modal reasoning ($ l = 1, 2$) and multi-modal fusion ($ l = 3$) respectively.

\begin{figure*}[t]
\centering
\includegraphics[width=0.82\linewidth]{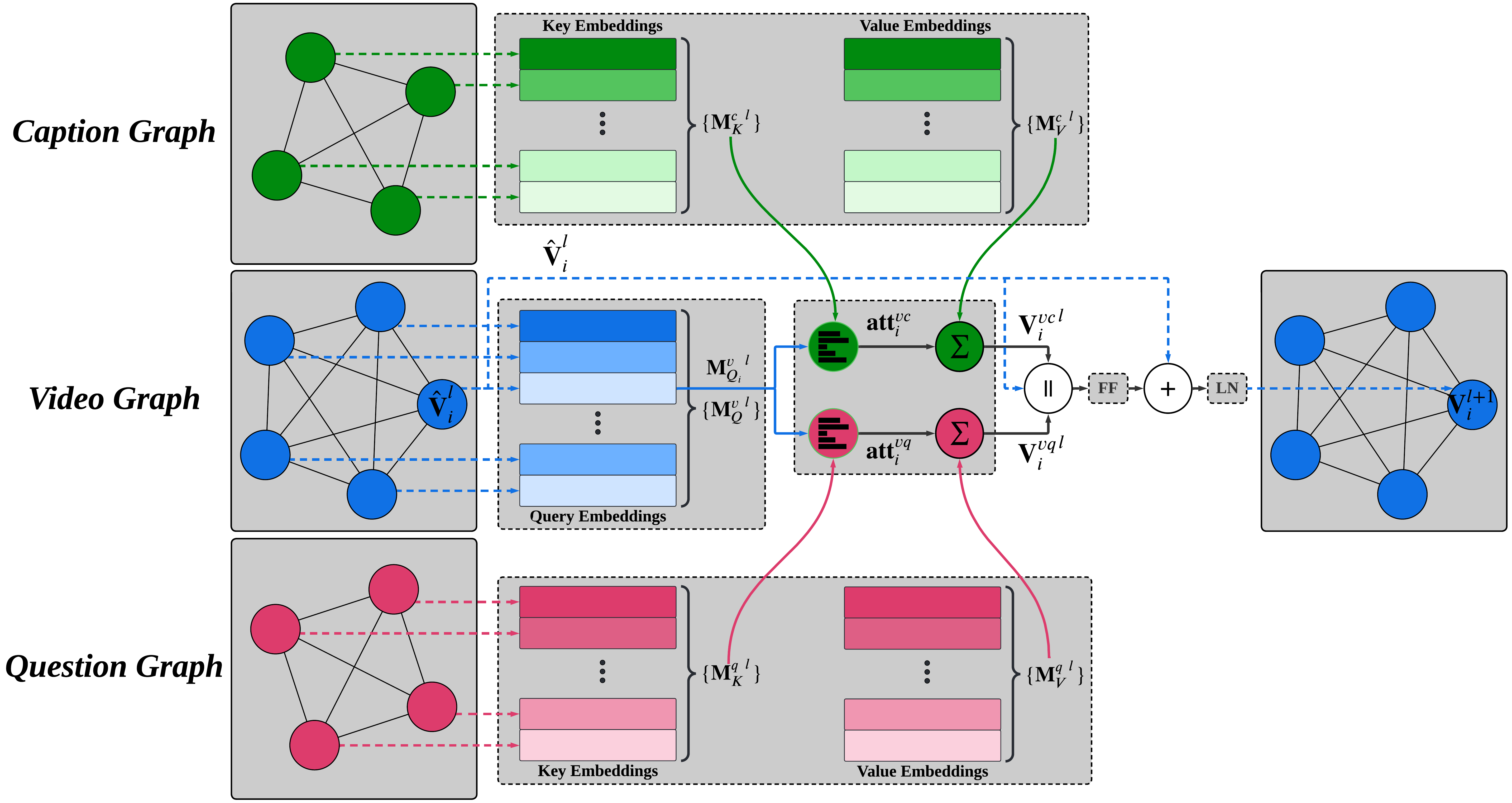}
\caption{An illustration of the cross-modal reasoning module.
For simplicity, the cross-modal reasoning module in \textit{video graph} between layer $l$ and $l + 1$ is presented, and the query node ${\bV}_{i}^{l}$ is taken as an example for demonstration.
Green, blue, and red arrows between modules corresponds to the information flow of caption, video, and question modalities respectively.
Dotted arrows represent the transmission of node features.
Different shades of color indicate representations of different nodes.
$\mbox{FF}$ and $\mbox{LN}$ denote feed-forward network and layer normalization.}
\label{Fig::Cross-Modal}
\end{figure*}

\subsection{Cross-Modal Reasoning with Cross-Modal Attention Mechanism}
\label{sec:method-cross-reasoning}
A video footage tends to focus on part of the question rather than the whole, and tends to pay more attention on a few captions instead of assigning equal weight to all captions.
Same rule applies for a word in the question and a caption in the set of dense captions.
Therefore, to infer the correct answer, we need to fully understand the dense interactions between factors of different modalities.

As shown in Figure~\ref{Fig::Architecture}, between two intra-modal graph reasoning procedures\eat{ of each modality}, we propose \eat{to leverage }a cross-modal attention mechanism (CAM) in cross-module reasoning procedure to explicitly exploit the inter-modal relationships with two other modalities.
For instance, in the cross-modal reasoning of \textit{video graph}, CAM allows a factor from the video modality as a clue to determine the weights of factors from question and caption modalities to aggregate the relevant information.

Given a query and a set of key-value pairs, CAM computes the weighted sum of values based on the dot-product similarity of the query and keys.
With query, key and value denoted as a set of vectors (i.e., ${\mM}_{Q}$, ${\mM}_{K}$, and ${\mM}_{V}$), we formulate the cross-modal attention mechanism as
\begin{align}
\mbox{CAM} ({\mM}_{Q}, {\mM}_{K}, {\mM}_{V}) &= \mbox{softmax}(\frac{{\mM}_{Q} {{\mM}_{K}}^{T}}{\sqrt{d}}) {\mM}_{V}
\end{align}
in which ${\mM}_{K}$ and ${\mM}_{V}$ belong to the same modality which is different from that of ${\mM}_{Q}$, and $d$ indicates the dimensionality of the vectors.
%

With CAM, which is designed to better embed and capture information across multiple modalities, as the core module,
we build the cross-modal reasoning procedure, which is illustrated in Figure~\ref{Fig::Cross-Modal} and elaborated in Appendix.
%
In brief, the cross-modal reasoning is performed on \textit{caption graph}, \textit{video graph}, and \textit{question graph}
with the input node features ${\hat{\bC}}^{l}$, ${\hat{\bV}}^{l}$, and ${\hat{\bQ}}^{l}$ from $l$-th graph reasoning module. Then, it produces a set of CAM-based features for the three modalities and further obtains the updated node features ${\bC}^{l + 1} = \{{\bC}^{l + 1}_{i}: i \leq N_{c}, {\bC}^{l + 1}_{i}  \in {\mR}^{d_{C}} \} $, ${\bV}^{l + 1} = \{{\bV}^{l + 1}_{i}: i \leq N_{v}, {\bV}^{l + 1}_{i}  \in {\mR}^{d_{V}} \} $, and ${\bQ}^{l + 1} = \{{\bQ}^{l + 1}_{i}: i \leq N_{q}, {\bQ}^{l + 1}_{i}  \in {\mR}^{d_{Q}} \} $, which are forwarded to the next GNN layer for the next round of intra-modal graph reasoning.
%

%
\begin{figure}[t]
\centering
\includegraphics[width=0.99\linewidth]{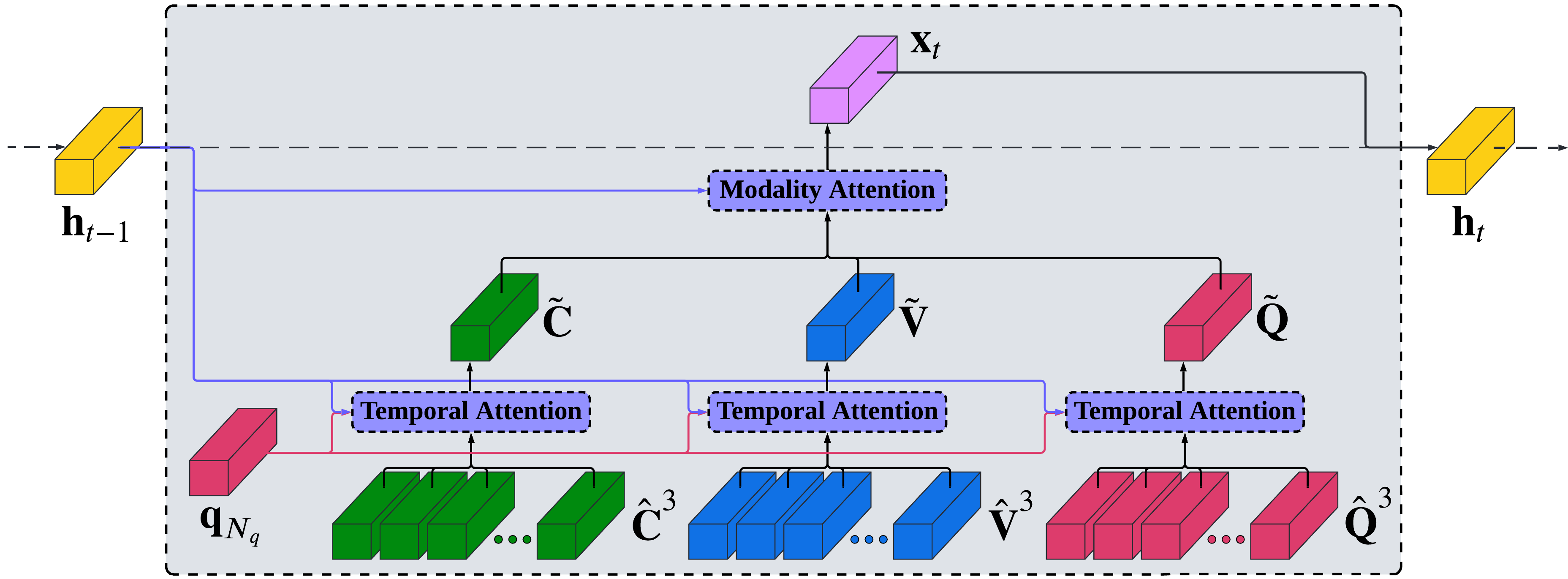}
\caption{An illustration of the question-guided self-adaptive multi-modal fusion module at the $t$-th step.
%
%
An LSTM controller with previous state ${\bh}_{t-1}$ attends to relevant caption features, visual features, and word features with question guidance, and combines them to obtain current state ${\bh}_{t}$.
%
}
\label{Fig::Cross-Fusion}
\end{figure}

\subsection{Question-Guided Self-Adaptive Multi-Modal Fusion}
\label{sec:method-fusion}
In multi-modal fusion procedure, to collect the question-oriented and event-correlated evidence from the three modalities,
%
%
we design a question-guided self-adaptive multi-modal fusion module, which performs multiple cycles of reasoning to selectively attend to multi-modal features through gradually refining the soft attention weights with the guidance of question.
%
%
Inspired by the multi-modal fusion module proposed by HME~\cite{fan2019heterogeneous}, we adopts multi-step reasoning with an LSTM controller.

%
The multi-modal fusion module takes the updated node features from the last GNN layer (i.e., caption features ${\hat{\bC}}^{3}$, visual features ${\hat{\bV}}^{3}$, and word features ${\hat{\bQ}}^{3}$) and the last hidden state of question (i.e., ${\bq}_{N_{q}}$) as the inputs.
The module deployed at each reasoning step is illustrated in Figure~\ref{Fig::Cross-Fusion}.
At each reasoning step \zp{$t$}, the module employs temporal attention mechanism to attend to different parts of the caption, visual, and word features independently, and collects the question-oriented information from each modality.
%
%
The attended features (i.e., $\tilde{\bC}$, $\tilde{\bV}$, and $\tilde{\bQ}$) are further incorporated together with learned modality weights to gather the event-correlated information across the modalities and produce representation ${\bh}_{t}$ of the current state, which is forwarded to the reasoning of next step.
By iteratively performing the reasoning step $N_{r}$ times, we obtain the final representation of the multi-modal fusion ${\bh}_{N_{r}}$.
Finally, we apply standard temporal attention~\cite{jang2017tgif} on the contextualized caption features ${\bC}^{1}$ and visual features ${\bV}^{1}$, and concatenate with ${\bh}_{N_{r}}$ to obtain the final representation ${\bs}_{a}$ for answer prediction.
The computation procedure and formulas in detail are available in Appendix.

\subsection{Answer Prediction}
\label{sec:method-question}
%
Based on the final representation ${\bs}_{a}$ from the fusion module, we predict the final answer of the VideoQA tasks.
We aim at solving three different types of VideoQA, same as \cite{jang2017tgif}: \textit{open-ended words}, \textit{open-ended numbers} and \textit{multiple-choice}.
Given the input video $v$ and the corresponding question $q$, the open-ended tasks is to infer an answer $\hat{a}$ from a pre-defined answer set $\mA$ of size $C$ that matches the ground-truth ${a}^{\star} \in {\mA}$, while the multiple-choice task is to select the correct answer $\hat{a}$ out of the candidate set $\{{a}_{i} \}_{i = 1}^{K}$.
%
With the representation ${\bs}_{a}$,
for each
of the three tasks, we train and evaluate a separate model with different predicting functions and training losses.
Detailed descriptions of the three predictors with training and inference settings can be found in Appendix.

%
\begin{table}[t]
\centering
\caption{Comparison on TGIF-QA in terms of MSE ($\downarrow$) for {\tcount} and Accuracy (\%, $\uparrow$) for others.}
\label{Tab::Res-TGIF-QA}
\resizebox{0.99\columnwidth}{!}{
\begin{tabular}{c|cccc}
Methods & \begin{tabular}[c]{@{}c@{}}{\tframeqa} ($\uparrow$)\\ \end{tabular} & \begin{tabular}[c]{@{}c@{}}{\tcount} ($\downarrow$)\\ \end{tabular} & \begin{tabular}[c]{@{}c@{}}{\taction} ($\uparrow$)\\ \end{tabular} & \begin{tabular}[c]{@{}c@{}}{\ttrans} ($\uparrow$)\\ \end{tabular} \\
\midrule
ST-VQA-Sp~\cite{jang2017tgif} & 45.5 & 4.28 & 57.3 & 63.7 \\
ST-VQA-Tp~\cite{jang2017tgif} & 49.3 & 4.40 & 60.8 & 67.1 \\
ST-VQA-SpTp~\cite{jang2017tgif} & 47.8 & 4.56 & 57.0 & 59.6 \\
ST-VQA$\star$~\cite{jang2019video} & 52.0 & 4.22 & 73.5 & 79.7 \\
Co-Mem~\cite{gao2018motion} & 51.5 & 4.10 & 68.2 & 74.3 \\
PSAC~\cite{li2019beyond} & 55.7 & 4.27 & 70.4 & 76.9 \\
HME~\cite{fan2019heterogeneous} & 53.8 & 4.10 & 73.9 & 77.8 \\
HGA~\cite{jiang2020reasoning} & 55.1 & 4.09 & 75.4 & 81.0 \\
\cmidrule{1-5}
Ours & \textbf{55.3} & \textbf{4.18} & \textbf{75.8} & \textbf{81.2}
\end{tabular}
}
\end{table}

\section{Performance Evaluation}
In this section, we evaluate our proposed model on two widely used large-scale VideoQA benchmarks and discuss the experimental results quantitatively and qualitatively.

\subsection{Benchmark Datasets}
\label{sec:benchmark-datasets}
\para{TGIF-QA} TGIF-QA~\cite{jang2017tgif} is the most commonly used benchmark for VideoQA with 165K QA pairs from 72 animated GIFs.
It includes four different VideoQA tasks:
(1) {\tframeqa}: An \textit{open-ended words} task, in which the question can be answered based on a single video frame.
(2) \tcount: An \textit{open-ended numbers} task, asking the number of repetition of a given action.
(3) \taction: A 5-option \textit{multiple-choice} task, which aims to recognize a repeated action given its specific times.
(4) \ttrans: A 5-option \textit{multiple-choice} task, asking the transition happened between two states.

\para{MSVD-QA} MSVD-QA~\cite{xu2017video} is generated automatically based on Microsoft Research Video Description Corpus~\cite{chen2011collecting}.
It contains 50K QA pairs associated with 1,970 video clips and consists of five different types of questions, including \twhat, \twho, \thow, {\twhen} and \twhere.
All questions are \textit{open-ended words} tasks, where the answer is predicted from a pre-defined answer set of size 1,000.

\subsection{Experimental Setup}
\label{sec:experimental-setup}
%
\para{Implementation details}
We employed the pre-trained ResNet-152 and VGG to obtain the appearance features for TGIF-QA and MSVD-QA respectively
%
%
and the pre-trained C3D to extract the motion features.
%
%
$d_{c}$, $d_{a}$, $d_{m}$, $d_{v}$, and $d_{q}$ are set to 512, 2,048, 4,096, 4,096, and 300, respectively.
%
%
${d_{V}}$, ${d_{Q}}$, and ${d_{C}}$ are all set to 512.
The dimensions of ${\bh}_{t}$ and ${\bs}_{a}$ are set to 512 and 1,536 respectively.
%
%
The proposed model was trained with Adam optimizer~\cite{kingma2014adam} with an initial learning rate $10^{-4}$ and a batch size 64.
Our code was implemented in PyTorch~\cite{paszke2017automatic} and run with NVIDIA Tesla P100 GPUs.

\para{Evaluation metrics}
For both \textit{open-ended words} and \textit{multiple-choice} tasks, we adopted the classification accuracy (i.e., Accuracy) as the evaluation metric.
For \textit{open-ended number} task, as all the answers belong to 11 possible integer values ranging from 0 to 10, we evaluated the model in terms of the mean square error (i.e., MSE) between the predicted integer and the ground-truth.
Note that the model delivers better performance with higher classification accuracy while lower mean square error.

\para{Baselines}
Our model was compared with some typical and state-of-the-art methods.
%
For TGIF-QA dataset, we compared {\Ours} with ST-VQAs~\cite{jang2017tgif}, ST-VQA$\star$~\cite{jang2019video}, Co-Mem~\cite{gao2018motion}, HME~\cite{fan2019heterogeneous}, PSAC~\cite{li2019beyond}, HGA~\cite{jiang2020reasoning} and HCRN~\cite{le2020hierarchical}.
For MSVD-QA dataset, in addition to ST-VQA, Co-Mem, HME and HGA, we also compared against E-VQA~\cite{zeng2017leveraging}, E-MN~\cite{zeng2017leveraging} and AMU~\cite{xu2017video}.
\subsection{Experimental Results}
We compare our proposed model with the aforementioned baselines.
The results on TGIF-QA and MSVD-QA datasets are shown in Tables~\ref{Tab::Res-TGIF-QA} and~\ref{Tab::Res-MSVD-QA},
and we can make the following observations.

\para{State-of-the art performance especially on event-related tasks}
{\Ours} outperforms most baselines and performs comparably with state-of-the-art methods (e.g., HGA~\cite{jiang2020reasoning} and HCRN~\cite{le2020hierarchical}) for both datasets.
Notably, {\Ours} delivers an accuracy of $75.8\%$ on the \textit{Action} task of TGIF-QA, excelling the second best by $0.4$ percentage points;
On MSVD-QA, it performs the best on {\twhat} task and the union of all tasks (denoted as {\tall} in Table~\ref{Tab::Res-MSVD-QA}).
The results suggest that leveraging event-related information (i.e., dense video captions) is beneficial for the cases where obvious event information (e.g., actions) is involved and multiple events occur concurrently or successively (e.g., state transitions),
and clearly demonstrate the effectiveness and feasibility of the proposed {\Ours}.

\para{Advantages of modality-aware graph networks}
Comparing to HGA~\cite{jiang2020reasoning}, which also uses GCNs for the modeling of intra-modal relationships, {\Ours} performs better on most tasks and question types.
The reason may be that HGA uses one graph, ignoring the pattern differences between visual and linguistic modalities and leading to semantic bias, while {\Ours} employ modality-aware graphs that model heterogeneous intra-modal relationships and revealing multi-modal features.

\para{Gains of question guidance}
{\Ours} consistently outperforms HME~\cite{fan2019heterogeneous}, which employs similar multi-modal fusion but lacks question guidance.
This suggests that gathering question-oriented information is crucial for a comprehensive understanding of cross-modal semantics.
An ablation study is conducted to further justify the superiority of question guidance in multi-modal fusion.

\para{Superiority of GCNs to RNNs}
EC-GNNs perform consistently better than E-VQA~\cite{zeng2017leveraging}, a typical RNNs-based VideoQA method.
It implies that for VideoQA tasks, GCNs learn better long-term dependencies than RNNs, especially when complex semantics are involved.
\begin{table}[t]
\centering
\caption{Comparison on MSVD-QA in terms of Accuracy (\%, $\uparrow$).}
\label{Tab::Res-MSVD-QA}
\resizebox{0.99\columnwidth}{!}{
\begin{tabular}{c|ccccc|c}
Methods & \begin{tabular}[c]{@{}c@{}} {\twhat} ($\uparrow$)\\ \end{tabular} & \begin{tabular}[c]{@{}c@{}}{\twho} ($\uparrow$)\\ \end{tabular} &
\begin{tabular}[c]{@{}c@{}}{\thow} ($\uparrow$)\\ \end{tabular} &
\begin{tabular}[c]{@{}c@{}}{\twhen} ($\uparrow$)\\ \end{tabular} & \begin{tabular}[c]{@{}c@{}}{\twhere} ($\uparrow$)\\ \end{tabular} &
\begin{tabular}[c]{@{}c@{}}{\tall} ($\uparrow$)\\ \end{tabular} \\
\midrule
E-VQA~\cite{zeng2017leveraging} & 9.7 & 42.2 & \textbf{83.8} & 72.4 & \textbf{53.6} & 23.3 \\
E-MN~\cite{zeng2017leveraging} & 12.9 & 46.5 & 80.3 & 70.7 & 50.0 & 26.7 \\
AMU~\cite{xu2017video} & 20.6 & 47.5 & 83.5 & 72.4 & \textbf{53.6} & 32.0 \\
ST-VQA~\cite{jang2017tgif} & 18.1 & 50.0 & \textbf{83.8} & 72.4 & 28.6 & 31.3 \\
Co-Mem~\cite{gao2018motion} & 19.6 & 48.7 & 81.6 & \textbf{74.1} & 31.7 & 31.7 \\
HME~\cite{fan2019heterogeneous} & 22.4 & 50.1 & 73.0 & 70.7 & 42.9 & 33.7 \\
HGA~\cite{jiang2020reasoning} & 23.5 & \textbf{50.4} & 83.0 & 72.4 & 46.4 & 34.7 \\
\cmidrule{1-7}
Ours & \textbf{24.0} & 49.9 & 79.7 & 70.7 & 50.0 & \textbf{34.8}
\end{tabular}
}
\end{table}
\begin{figure}[h!]
\centering
\includegraphics[width=0.99\linewidth]{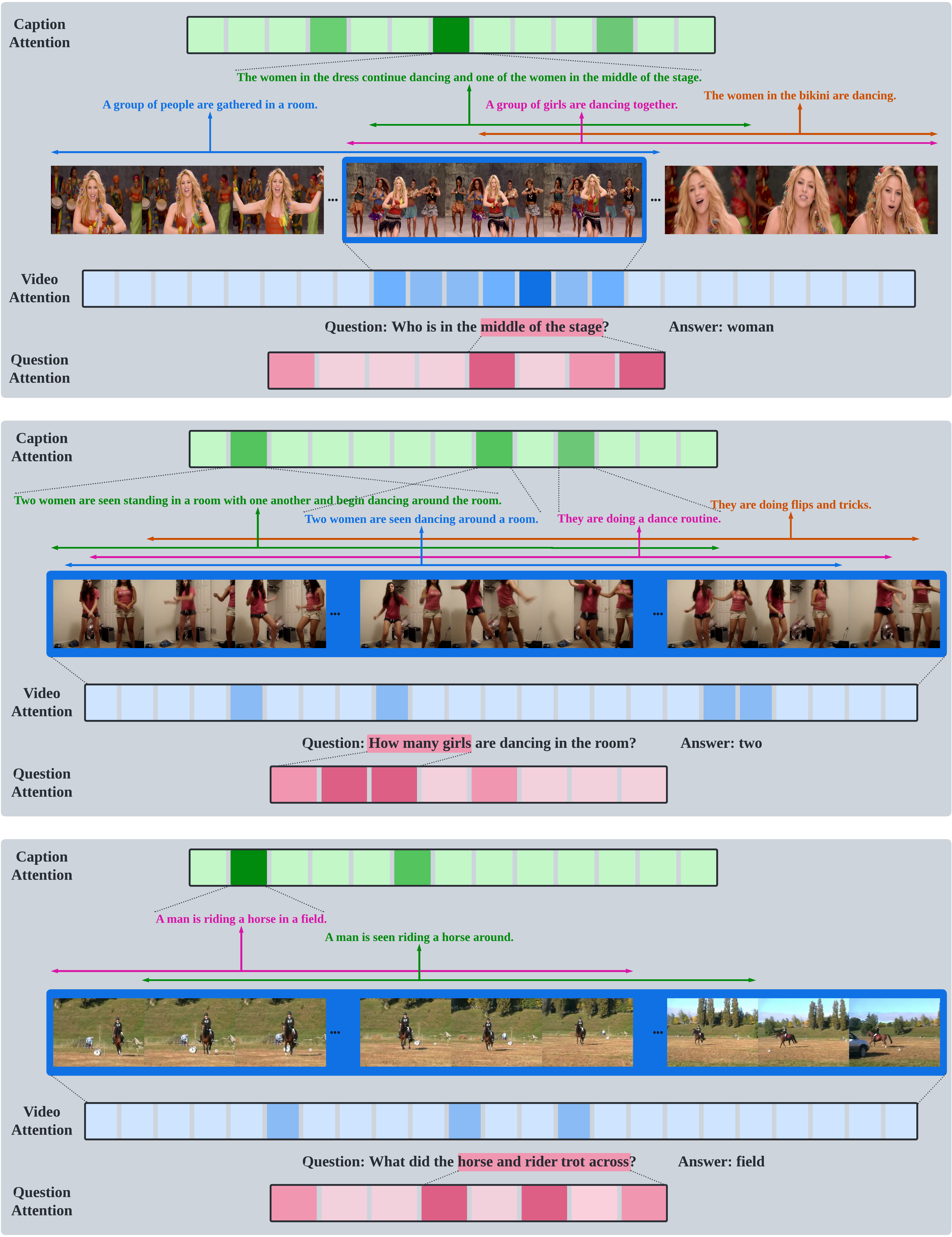}
\caption{Visualization of learned attention weights over the caption (in green), visual (in blue), and word (in red) features. Dark/light color denotes higher/lower values.}
\label{Fig::Vis}
\end{figure}

\subsection{Visualization}
%
To demonstrate the efficacy of the proposed model,
we use Figure~\ref{Fig::Vis} to visualize the learned attention weights\footnote[2]{Please refer to Appendix
for the definitions and equations of the learned attention weights.}
over caption, visual, and word features in the question-guided self-adaptive multi-modal fusion.
%
%
%

%

The first sample shows that the proposed model can effectively focus on the most relevant frames, words and captions for a given question. 
This example is from the MSVD-QA dataset.
From the point view of the video as a whole, it is difficult to know \textit{who is in the middle of the stage}, especially when close-up shots take up most of the video.
Furthermore, the model needs to find out the most important ones from a large number of dense video captions, which helps to infer the correct answer.
%
%
Our model subtly pays more attention to a few overall shots in the middle of the video, where a woman leads the dance in the center of the stage while others dance around.
Correspondingly, in the caption modality, the model assigns a higher weight to the caption \textit{the women in the dress continue dancing and one of the women in the middle of the stage} while reducing other weights.

In the latter two samples, the model is able to directly infer the correct answer by referring to some correlated captions.
In addition to providing some visually intuitive information, such as the presence of two women in the video of the second sample,
the captions can also incorporate some commonsense knowledge that may not be represented by an end-to-end model for VideoQA datasets with limited sample diversity~\cite{khademi2020multimodal}.
For instance, in the third sample, the caption \textit{a man is riding a horse in a field} can naturally associate the \textit{horse riding} with \textit{field}, which provides exactly where \textit{the horse and rider trot across}.
Nonetheless, without assistance from dense video captions, most other models have to recognize the environment based on visual features, which is quite challenging due to the ambiguities contained in video frames.

\begin{table}[t]
\centering
\caption{Ablation study on TGIF-QA. MSE ($\downarrow$) for {\tcount} and Accuracy (\%, $\uparrow$) for others.
(Vid: Video Modality; Cap: Caption Modality; CMR: Cross-Modal Reasoning Module; MMF: Multi-Modal Fusion module from \cite{fan2019heterogeneous}; Q-MMF: Question-guided self-adaptive Mutli-Modal Fusion module.)
}
\label{Tab::Res-Abl-TGIF-QA}
\resizebox{0.99\columnwidth}{!}{
\begin{tabular}{c|c|c|c|c|c|cccc}
Case \# & Vid & Cap & CMR & MMF & Q-MMF & \begin{tabular}[c]{@{}c@{}}{\tframeqa} ($\uparrow$)\\ \end{tabular} & \begin{tabular}[c]{@{}c@{}}{\tcount} ($\downarrow$)\\ \end{tabular} & \begin{tabular}[c]{@{}c@{}}{\taction} ($\uparrow$)\\ \end{tabular} & \begin{tabular}[c]{@{}c@{}}{\ttrans} ($\uparrow$)\\ \end{tabular} \\
\midrule
1 & \ding{51} &  & \ding{51} &  & \ding{51} & 52.1 & 4.33 & 73.4 & 77.5 \\
2 & \ding{51} & \ding{51} &  &  & \ding{51} & 53.3 & 4.28 & 74.0 & 76.8 \\
3 & \ding{51} & \ding{51} & \ding{51} &  &  & 52.6 & 4.32 & 74.7 & 78.5 \\
4 & \ding{51} & \ding{51} & \ding{51} & \ding{51} &  & 54.1 & \textbf{4.15} & 75.3 & 80.8 \\
\cmidrule{1-10}
ALL & \ding{51} & \ding{51} & \ding{51} &  & \ding{51} & \textbf{55.3} & 4.18 & \textbf{75.8} & \textbf{81.2}
\end{tabular}
}
\end{table}

\subsection{Ablation Study}
To validate the contribution of each component in the proposed {\Ours}, several ablation experiments were conducted on TGIF-QA.
The following observations are made based on Table~\ref{Tab::Res-Abl-TGIF-QA}.

\para{Benifits of generating an event-correlated modality}
{\Ours} incorporates dense video captions as a new modality to perform reasoning of VideoQA, which leads to better performance.
%
Compared with Case-ALL, which learns the full model, the accuracy of the \textit{Action} task decreases by $2.4\%$ in Case-1, where the caption modality is not modeled.
%
This suggests that dense video captions can provide valuable evidence for answer reasoning, especially when the videos and questions contain complex semantics.
Similar trends can be observed in \textit{FrameQA}, \textit{Count} and \textit{Trans} tasks.

\para{Effectiveness of cross-modal reasoning}
The model in Case-2, without the cross-modal reasoning module, gets $1.8\%$ accuracy degradation of the \textit{Action} task compared with Case-ALL.
Degraded performance of this ablated model is also observed on the other three tasks.
This demonstrates the benefits of cross-modal reasoning in the modeling inter-modal relationships, which can aggregate relevant information across different modalities.

\para{Gains of question guidance in multi-modal fusion}
The leveraging of multi-modal fusion delivers a performance gain over Case-3, where only the self-attention pooling and bi-linear fusion are applied for feature fusion~\cite{jiang2020reasoning}.
For instance, for the \textit{Action} task, Case-4 brings up improvements of $0.6\%$ on accuracy,
and further adopting question-guided self-adaptive multi-modal fusion in Case-ALL achieves an additional gain of $0.5\%$.
Similar trends can be observed in \textit{FrameQA} and \textit{Trans} tasks.
In brief, collecting the question-oriented and event-correlated evidence facilitates a comprehensive understanding of cross-modal semantics and outperforms simple multi-modal fusion.

\section{Conclusions}
By introducing \textbf{E}vent-\textbf{C}orrelated \textbf{G}raph \textbf{N}eural \textbf{N}etwork\textbf{s} (\textbf{EC-GNNs}), we propose a unified end-to-end trainable model for Video Question Answering (VideoQA).
%
Notably, this paper clarifies how event-correlated information can be extracted as a new modality and integrated into VideoQA inference.
Furthermore, we solve the VideoQA task through a multi-modal graph consisting of three modality-aware graph convolution networks that perform cross-modal reasoning over three modalities (i.e., the caption, the video, and the question).
Comprehensive experiments have been conducted on two widely-used VideoQA benchmarks, and extensive experimental results show the superiority of introducing the event-based caption modality and the effectiveness of the proposed model equipped with the cross-modal attention mechanism and the question guidance.

\clearpage

\bibliography{EC-GNNs-VideoQA-AAAI-2022}

\clearpage

\appendix
\onecolumn
\section{Appendix}
\subsection{Methodological Details}

\subsubsection{Representing Video and Question Modalities} \label{sup:method-representation}
%
In addition to the generated caption modality described in the paper, the video modality and question modality are also properly represented and utilized in the proposed model.

\para{Representing video modality}
In order to take into account the appearance and motion information involved in video contents, we represent a video at both frame-level and shot-level.
Specifically, to obtain appearance features, we use the 2D ConvNets (i.e., ResNet-152~\cite{he2016deep} and VGG~\cite{simonyan2014very}) pre-trained on the ImageNet 2012 classification dataset~\cite{russakovsky2015imagenet}.
For shot-level motion features, we leverage 3D ConvNets (i.e., C3D~\cite{tran2015learning}) pre-trained on the Sport1M dataset~\cite{karpathy2014large}.
Then, the video is represented as two feature views, appearance features ${\bF}^{a} = \{{\bbf}_{i}^{a}: i \leq N_{v}, {\bbf}_{i}^{a} \in {\mR}^{d_{a}}\}$, and motion features ${\bF}^{m} = \{{\bbf}_{i}^{m}: i \leq N_{v}, {\bbf}_{i}^{m} \in {\mR}^{d_{m}}\}$, where $N_{v}$ is number of frames and $d_{a}$, $d_{m}$ are dimensionalities of the two feature views.
We project the concatenation of the two features into a common visual space by two fully-connected layers, to obtain a joint representation of the video ${\bF}^{v} = \{{\bbf}_{i}^{v}: i \leq N_{v}, {\bbf}_{i}^{v} \in {\mR}^{d_{v}}\}$, where $d_{v}$ represents the dimensionality of visual features.
%

\para{Representing question modality}
For a given question, we represent each word as a fixed-length vector initialized with the GloVe $300D$ word embedding~\cite{pennington2014glove} pre-trained on the Common Crawl dataset.
The question is then denoted as a sequence of word embeddings ${\bF}^{q} = \{{\bbf}_{i}^{q}: i \leq N_{q}, {\bbf}_{i}^{q} \in {\mR}^{d_{q}}\}$, in which $N_{q}$ represents the number of words and $d_{q}$ equals 300.
%
%

\subsubsection{Intra-Modal Graph Reasoning}
\label{sup:method-reasoning}

The intra-modal graph reasoning procedures in these three graphs share the common operations but differ in their node representations.
%
%
Thus we take the \textit{video graph} as an instance to illustrate the operations of graph reasoning.

We first project the node features ${\bV}^{l}$ into an interaction space by a non-linear transformation operation $\phi(.)$.
Then we calculate the dot-product similarity~\cite{wang2018videos}, which measures the semantic correlations among nodes, and employ softmax function on each row of the matrix to obtain the normalized adjacency matrix.
\begin{align}
{\bG}_{V}^{l} &= \mbox{softmax}(\phi({\bV}^{l}) \phi({\bV}^{l})^{T})
\end{align}
where ${\bG}_{V}^{l} \in {\mR}^{N_{v} \times N{v}}$ represents the adjacency matrix and ${{\bG}_{V}}_{i,j}^{l}$ indicates the weight between ${\bV}_{i}^{l}$ and ${\bV}_{j}^{l}$.

To perform intra-modal reasoning on the graph, we apply the Graph Convolutional Networks (GCNs) proposed in \cite{kipf2016semi}, which allows us to update each node based on its neighbors and the corresponding weights.
Given the node features ${\bV}^{l}$ and weights specified by the adjacency matrix ${\bG}_{V}^{l}$, the nodes are updated by a linear transformation of aggregated excitation of its neighbors and itself.
Formally, we represent the graph convolution on layer $l$ of \textit{video graph} as
\begin{align}
{\hat{\bV}}^{l} &= \mbox{Relu}(\mbox{LayerNorm}({\bG}_{V}^{l} {\bV}^{l} {\bW}))
\end{align}
where ${\bW} \in {\mR}^{d_{V} \times d_{V}}$ is a learnable weight matrix and ${\hat{\bV}}^{l}= \{{\hat{\bV}}^{l}_{i}: i \leq N_{v}, {\hat{\bV}}^{l}_{i} \in {\mR}^{d_{V}}\}$ represents the updated node features.
Two non-linear functions including Layer Normalization~\cite{ba2016layer} and ReLU are applied after the graph convolution.

We conduct the above intra-modal graph reasoning on \textit{caption graph}, \textit{video graph}, and \textit{question graph} independently and obtain the updated node features ${\hat{\bC}}^{l}$, ${\hat{\bV}}^{l}$ and ${\hat{\bQ}}^{l}$ accordingly,
which are forwarded to cross-modal reasoning ($ l = 1, 2$) and multi-modal fusion ($ l = 3$) respectively.

\begin{figure*}[t]
\centering
\includegraphics[width=0.90\linewidth]{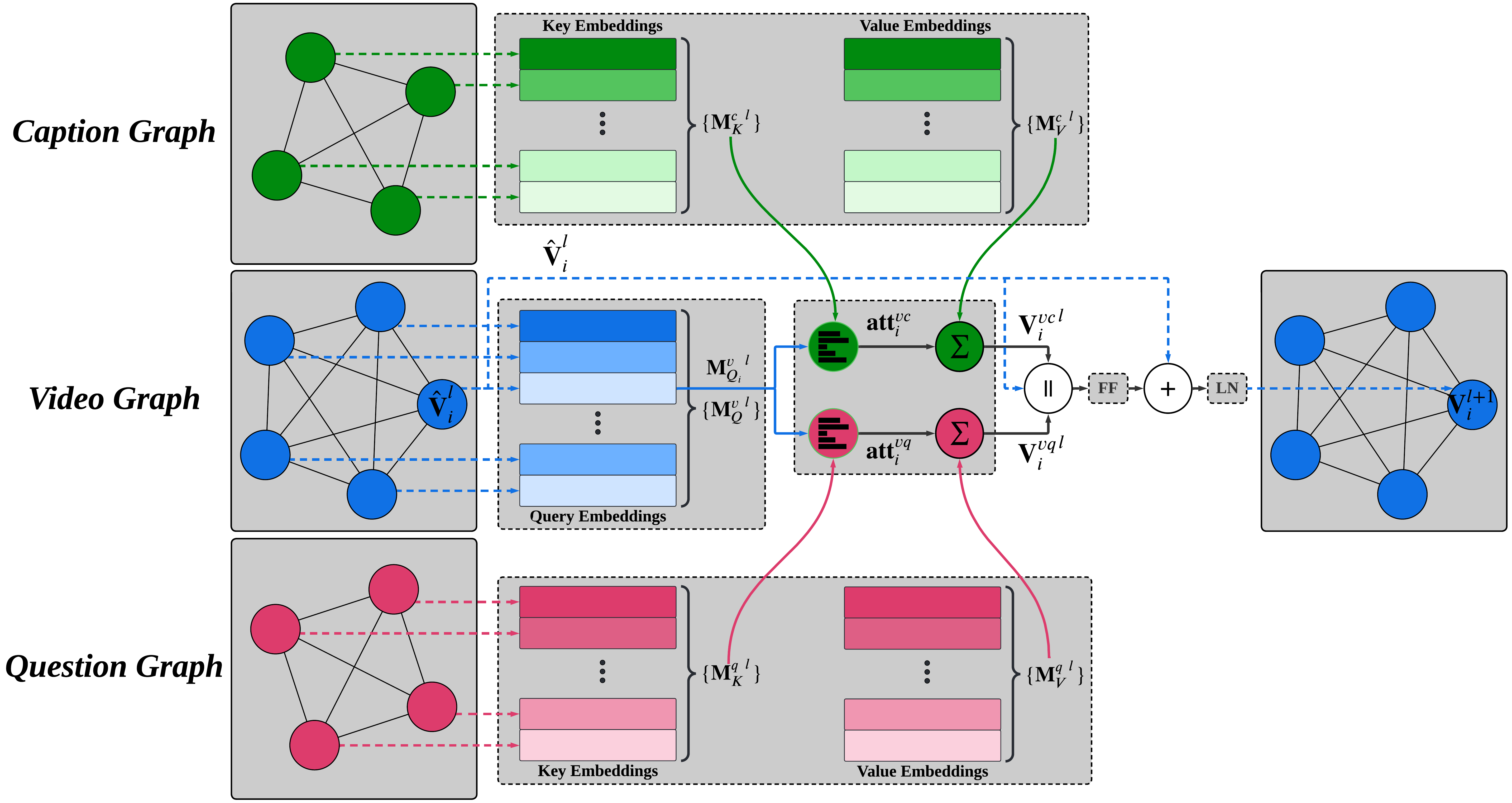}
\caption{An illustration of the cross-modal reasoning module.
For simplicity, the cross-modal reasoning module in \textit{video graph} between layer $l$ and $l + 1$ is presented, and the query node ${\bV}_{i}^{l}$ is taken as an example for demonstration.
Green, blue, and red arrows between modules corresponds to the information flow of caption, video, and question modalities respectively.
Dotted arrows represent the transmission of node features.
Different shades of color indicate representations of different nodes.
$\mbox{FF}$ and $\mbox{LN}$ denote feed-forward network and layer normalization.}
\label{Fig::sup-Cross-Modal}
\end{figure*}

\subsubsection{Operations of Cross-Modal Graph Reasoning}
\label{sup:method-cross-modal}
The cross-modal attention mechanism (CAM) is used to explicitly exploit the inter-modal relationships of one modality with two other modalities.
As same operations shared across the cross-modal reasoning procedures in three graphs,
in Figure~\ref{Fig::sup-Cross-Modal}, we give a description of cross-modal reasoning using \textit{video graph} as an example, in which the node features of \textit{video graph} are employed as query vectors, and node features of \textit{question graph} and \textit{caption graph} serve as key-value pairs.

We first linearly project the updated node features ${\hat{\bV}}^{l}$, ${\hat{\bQ}}^{l}$ and ${\hat{\bC}}^{l}$ into a transformed space.
\begin{align}
{{\bM}_{Q}^{v}}^{l} &= {\bW}_{Q}^{v} {\hat{\bV}}^{l} \\
{{\bM}_{K}^{q}}^{l}, {{\bM}_{V}^{q}}^{l} &= {\bW}_{K}^{q} {\hat{\bQ}}^{l}, {\bW}_{V}^{q} {\hat{\bQ}}^{l} \\
{{\bM}_{K}^{c}}^{l}, {{\bM}_{V}^{c}}^{l} &= {\bW}_{K}^{c} {\hat{\bC}}^{l}, {\bW}_{V}^{c} {\hat{\bC}}^{l}
\end{align}
where ${\bW}_{Q}^{v}$, $ {\bW}_{K}^{q}$, ${\bW}_{V}^{q}$, ${\bW}_{K}^{c}$ and ${\bW}_{V}^{c}$ are the learnable weight matrices.
${{\bM}_{Q}^{v}}^{l} = \{{{\bM}_{{Q}_{i}}^{v}}^{l}: i \leq N_{v}\}$ represents the transformed visual features in the query space.
${{\bM}_{K}^{q}}^{l} = \{{{\bM}_{{K}_{i}}^{q}}^{l}: i \leq N_{q}\}$ and ${{\bM}_{V}^{q}}^{l} = \{{{\bM}_{{V}_{i}}^{q}}^{l}: i \leq N_{q}\}$ represent the transformed word features in the key and value spaces.
${{\bM}_{K}^{c}}^{l} = \{{{\bM}_{{K}_{i}}^{c}}^{l}: i \leq N_{c}\}$ and ${{\bM}_{V}^{c}}^{l} = \{{{\bM}_{{V}_{i}}^{c}}^{l}: i \leq N_{c}\}$ represent the transformed caption features in the key and value spaces.
Then we apply the CAM to embed the visual information into the feature spaces of question words and dense captions respectively to collect the relevant information from question and caption modalities.
\begin{align}
{{\bV}^{vq}}^{l} &= \mbox{CAM}({{\bM}_{Q}^{v}}^{l}, {{\bM}_{K}^{q}}^{l}, {{\bM}_{V}^{q}}^{l}) \\
{{\bV}^{vc}}^{l} &= \mbox{CAM}({{\bM}_{Q}^{v}}^{l}, {{\bM}_{K}^{c}}^{l}, {{\bM}_{V}^{c}}^{l})
\end{align}
where ${{\bV}^{vq}}^{l} = \{{{\bV}_{i}^{vq}}^{l}: i \leq N_{v}\}$ and ${{\bV}^{vc}}^{l} = \{{{\bV}_{i}^{vc}}^{l}: i \leq N_{v}\}$ indicate the word features and caption features attended by the visual features (i.e., CAM-based features), which are combined with the node features ${\hat{\bV}}^{l}$ to obtain the updated visual features ${\bV}^{l + 1} = \{{\bV}^{l + 1}_{i}: i \leq N_{v}, {\bV}^{l + 1}_{i}\} $ after cross-modal reasoning.
\begin{align}
{\bV}^{l + 1} &= \mbox{LayerNorm}(\mbox{FF}([{{\bV}^{vq}}^{l} || {{\bV}^{vc}}^{l} || {\hat{\bV}}^{l}]) + {\hat{\bV}}^{l})
\end{align}
where $||$ denotes the concatenation operation and FF represents a feed-forward network. To achieve a better performance, a residual connection~\cite{he2016deep} is adopted for multiple layers stacking.

The above cross-modal reasoning is performed on \textit{video graph}, \textit{question graph} and \textit{caption graph} and obtains the updated node features ${\bV}^{l + 1}$, ${\bQ}^{l + 1}$ and ${\bC}^{l + 1}$, which are forwarded to the next GNN layer for the next round of intra-modal graph reasoning.

%
\begin{figure}[t]
\centering
\includegraphics[width=0.99\linewidth]{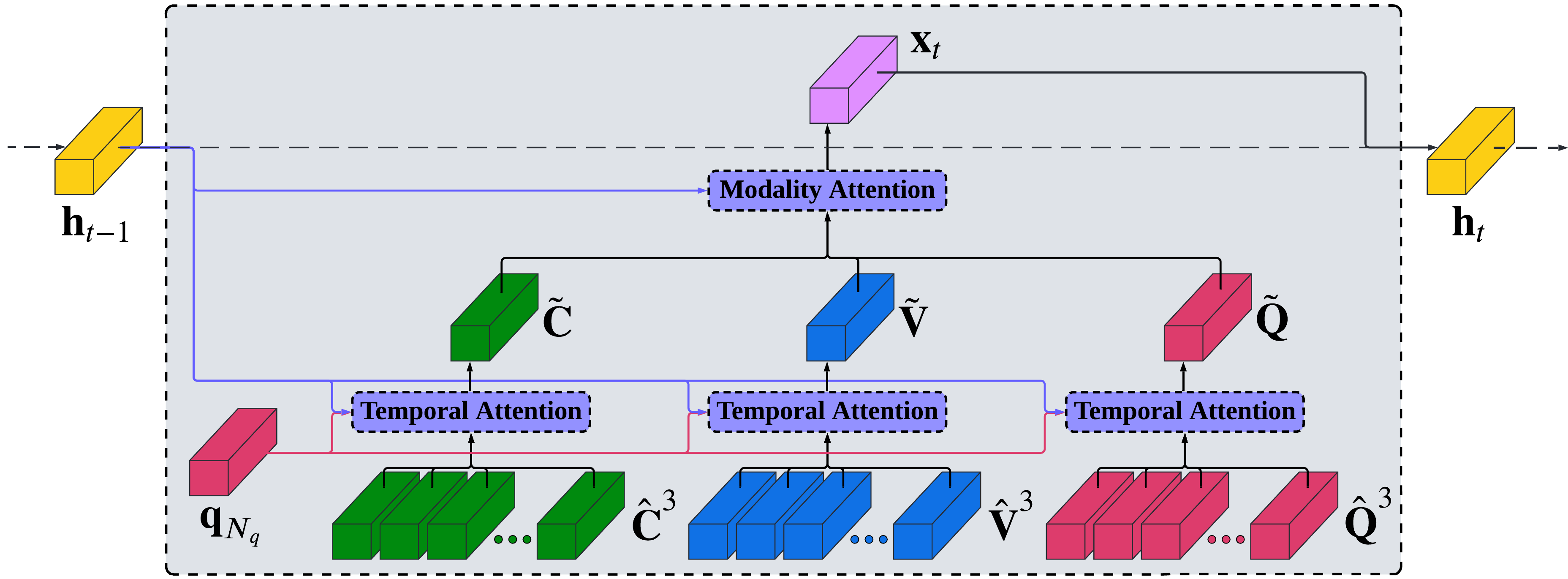}
\caption{An illustration of the question-guided self-adaptive multi-modal fusion module at the $t$-th step.
%
%
An LSTM controller with previous state ${\bh}_{t-1}$ attends to relevant visual features, word features and caption features with question guidance, and combines them to obtain current state ${\bh}_{t}$.
%
}
\label{Fig::sup-Cross-Fusion}
\end{figure}

\subsubsection{Calculating Question-Guided Self-Adaptive Multi-Modal Fusion}
\label{sup:method-fusion}
The multi-modal fusion module takes the updated node features from the last GNN layer (i.e., caption features ${\hat{\bC}}^{3}$, visual features ${\hat{\bV}}^{3}$, and word features ${\hat{\bQ}}^{3}$) and the last hidden state of question (i.e., ${\bq}_{N_{q}}$) as the inputs.
For the multi-step reasoning, inspired by the multi-modal fusion module in HME~\cite{fan2019heterogeneous},
we adopt an LSTM controller with its previous state denoted as ${\bh}_{t - 1}$.

At $t$-th step of reasoning, as shown in Figure~\ref{Fig::sup-Cross-Fusion}, by interacting with the last hidden state of question as well as the previous state of LSTM controller, the module first employs temporal attention mechanism to attend to different parts of the caption features, visual features, and word features, independently, and collects the question-oriented information from each modality.
\begin{align}
\label{eq:attc}
{\batt}^{c} &= \mbox{softmax}({\bW}^{c} \mbox{tanh}({\bW}_{q}^{c} {\bq}_{N_{q}} + {\bW}_{h}^{c} {\bh}_{t - 1} + {\bW}_{c}^{c} {\hat{\bC}}^{3} + {\bb}^{c})) \\
\label{eq:attv}
{\batt}^{v} &= \mbox{softmax}({\bW}^{v} \mbox{tanh}({\bW}_{q}^{v} {\bq}_{N_{q}} + {\bW}_{h}^{v} {\bh}_{t - 1} + {\bW}_{v}^{v} {\hat{\bV}}^{3} + {\bb}^{v})) \\
\label{eq:attq}
{\batt}^{q} &= \mbox{softmax}({\bW}^{q} \mbox{tanh}({\bW}_{q}^{q} {\bq}_{N_{q}} + {\bW}_{h}^{q} {\bh}_{t - 1} + {\bW}_{q}^{q} {\hat{\bQ}}^{3} + {\bb}^{q})) \\
\tilde{\bC} =& \sum_{i = 1}^{N_{c}} {\batt}_{i}^{c} {\hat{\bC}}_{i}^{3} \quad
\tilde{\bV} = \sum_{i = 1}^{N_{v}} {\batt}_{i}^{v} {\hat{\bV}}_{i}^{3} \quad
\tilde{\bQ} = \sum_{i = 1}^{N_{q}} {\batt}_{i}^{q} {\hat{\bQ}}_{i}^{3}
\end{align}
where ${\batt}^{c} \in {\mR}^{N_{c}}$, ${\batt}^{v} \in {\mR}^{N_{v}}$, and ${\batt}^{q} \in {\mR}^{N_{q}}$ indicate the normalized attention weights over the caption features ${\hat{\bC}}^{3}$, visual features ${\hat{\bV}}^{3}$, and word features ${\hat{\bQ}}^{3}$ respectively.
${\bW}^{c}$, ${\bW}^{v}$, ${\bW}^{q}$, ${\bW}_{q}^{c}$, ${\bW}_{q}^{v}$, ${\bW}_{q}^{q}$, ${\bW}_{h}^{c}$, ${\bW}_{h}^{v}$, ${\bW}_{h}^{q}$, ${\bW}_{c}^{c}$, ${\bW}_{v}^{v}$, ${\bW}_{q}^{q}$, ${\bb}^{c}$, ${\bb}^{v}$ and ${\bb}^{q}$ are the learnable weights.
$\tilde{\bC}$, $\tilde{\bV}$, and $\tilde{\bQ}$ represent the attended caption features, attended visual features, and attended word features respectively, which are then incorporated together with the learned modality weights ${\balpha} = \{ {\alpha}^{c}, {\alpha}^{v}, {\alpha}^{q} \}$ to gather the event-correlated information across the three modalities.
\begin{align}
{\balpha} &= \mbox{softmax}({\bW}^{\alpha} \mbox{tanh}([{\bW}_{c}^{\alpha} \tilde{\bC} || {\bW}_{v}^{\alpha} \tilde{\bV} || {\bW}_{q}^{\alpha} \tilde{\bQ}] + {\bW}_{h}^{\alpha} {\bh}_{t - 1} + {\bb}^{\alpha})) \\
{{\bx}_{t}} &= \mbox{tanh}({\alpha}^{c} {\bW}_{c}^{x} \tilde{\bC} + {\alpha}^{v} {\bW}_{v}^{x} \tilde{\bV} + {\alpha}^{q} {\bW}_{q}^{x} \tilde{\bQ} + {\bW}_{h}^{x} {\bh}_{t - 1} + {\bb}^{x})
\end{align}
where ${\bW}^{\alpha}$, ${\bW}_{c}^{\alpha}$, ${\bW}_{v}^{\alpha}$, ${\bW}_{q}^{\alpha}$, ${\bW}_{h}^{\alpha}$, ${\bW}_{c}^{x}$, ${\bW}_{v}^{x}$, ${\bW}_{q}^{x}$, ${\bW}_{h}^{x}$, ${\bb}^{\alpha}$ and ${\bb}^{x}$ are the learnable weights.
The output feature ${{\bx}_{t}}$ is forwarded to obtain the current state ${\bh}_{t}$ of the LSTM controller by ${\bh}_{t} = \mbox{LSTM}({{\bx}_{t}}, {\bh}_{t - 1})$.

We iteratively perform the above reasoning step to obtain the final representation of the multi-modal fusion.
After $N_{r}$ steps, the state LSTM controller ${\bh}_{N_{r}}$ captures the question-oriented and event-correlated information across the three modalities.
We set the value of $N_{r}$ to 3, referring to that in HME~\cite{fan2019heterogeneous}.
Following \cite{jang2017tgif}, we apply the standard temporal attention on the contextualized visual features ${\bV}^{1}$ and caption features ${\bC}^{1}$, and concatenate with ${\bh}_{N_{r}}$ to obtain the final representation ${\bs}_{a}$.

\subsubsection{Answer Prediction from Final Representation}
\label{sup:method-prediction}

In this section, we introduce how to predict the correct answer based on the final representation ${\bs}_{a}$.
Same as \cite{jang2017tgif}, we are dealing with the VideoQA tasks in three different types, \textit{open-ended words}, \textit{open-ended numbers} and \textit{multiple-choice}.
Given the input video $v$ and the corresponding question $q$, the open-ended tasks is to infer an answer $\hat{a}$ from a pre-defined answer set $\mA$ of size $C$ that matches the ground-truth ${a}^{\star} \in {\mA}$, while the multiple-choice task aims to select the correct answer $\hat{a}$ out of the candidate set $\{{a}_{i} \}_{i = 1}^{K}$.
Note that a separate model is trained and evaluated for each type of class.

\para{Open-ended word task}
The \textit{open-ended words} is treated as a classification problem.
We apply a linear layer and a softmax function upon the representation ${\bs}_{a}$ to generate the probabilities ${\bp}$ for all answers in the pre-defined answer set $\mA$.
\begin{align}
{\bp} &= \mbox{softmax} ({\bW}_{a}^{w} {\bs}_{a} + {\bb}^{w})
\end{align}
in which ${\bW}_{a}^{w}$ and ${\bb}^{w}$ are the learnable weights.
The cross-entropy loss is adopted to train the model, and the predicted answer is represented as ${\hat{a}} = \mbox{argmax}_{{\mA}}({\bp})$ in testing phase.

\para{Open-ended number task}
We consider the \textit{open-ended numbers} as a linear regression problem, which takes the representation ${\bs}_{a}$ and outputs a rounded integer value as the predicted answer.
\begin{align}
{\hat{a}} &= \mbox{round} ({\bW}_{a}^{n} {\bs}_{a} + {\bb}^{n})
\end{align}
where ${\bW}_{a}^{n}$ and ${\bb}^{n}$ are the learnable weights.
The ${l}_{2}$ loss is leveraged to minimize the gap between the predicted integer and the ground-truth.
Note that $\mA$ is an integer-valued answer set (i.e., 0-10) for the \textit{open-ended numbers}.

\para{Multiple choice task}
For \textit{multiple-choice}, we concatenate the question $q$ with each answer in the candidate set $\{{a}_{i} \}_{i = 1}^{K}$ to obtain $K$ word sequences,
%
%
which are forward to the model individually to get $K$ final representations $\{ {\bs}_{a} \}_{i = 1}^K$.
Then we use a linear function to transfer the $K$ representations to $K$ scores ${\bs} = \{ s^{p}, s_{1}^{n}, s_{2}^{n}, ..., s_{K - 1}^{n} \}$.
The answer with the highest score is selected as the predicted one.
We optimize the model by maximizing the hinge loss ${L_{hinge}}$ between the score for the correct answer $s^{p}$ and the scores for the incorrect answers $ \{ s_{i}^{n} \}_{i = 1}^{K - 1}$.
%
%
\begin{align}
{L_{hinge}} &= \sum_{i = 1}^{K - 1} \mbox{max}(0, 1 + s_{i}^{n} - s^{p})
\end{align}
%
%


\end{document}